%% file: main.tex
\documentclass[10pt,twocolumn,letterpaper]{article}

\usepackage{cvpr}              % To produce the CAMERA-READY version
\input{preamble}

\definecolor{cvprblue}{rgb}{0.21,0.49,0.74}
\usepackage[pagebackref,breaklinks,colorlinks,citecolor=cvprblue]{hyperref}

\def\paperID{*****} % *** Enter the Paper ID here
\def\confName{CVPR}
\def\confYear{2024}

\title{USat: A Unified Self-Supervised Encoder for Multi-Sensor Satellite Imagery}

\author{Jeremy Irvin\thanks{Equal contribution.}, Lucas Tao\footnotemark[1], Joanne Zhou, Yuntao Ma, Langston Nashold, Benjamin Liu, Andrew Y. Ng\\
Stanford University\\
{\tt\small \{jirvin16,lucastao,joannezhou,yunt.ma,lnashold,bencliu,ayn\}@stanford.edu}
}

\begin{document}
\maketitle
\input{sec/0_abstract}
\vspace{-2em}
\input{sec/1_introduction}
\input{sec/2_methods}

\input{sec/3_experiments}

\input{sec/4_discussion}
\input{sec/5_conclusion}
{
    \small
    \bibliographystyle{ieeenat_fullname}
    \bibliography{main}
}

% % WARNING: do not forget to delete the supplementary pages from your submission 
\input{sec/6_supplementary}

\end{document}

%% file: preamble.tex
\usepackage[dvipsnames]{xcolor}

\usepackage{graphicx}
\usepackage{multirow}
\usepackage{amsmath}
\usepackage{amssymb}

%% file: sec/0_abstract.tex
\begin{abstract}
Large, self-supervised vision models have led to substantial advancements for automatically interpreting natural images. Recent works have begun tailoring these methods to remote sensing data which has rich structure with multi-sensor, multi-spectral, and temporal information providing massive amounts of self-labeled data that can be used for self-supervised pre-training. In this work, we develop a new encoder architecture called USat that can input multi-spectral data from multiple sensors for self-supervised pre-training. USat is a vision transformer with modified patch projection layers and positional encodings to model spectral bands with varying spatial scales from multiple sensors. We integrate USat into a Masked Autoencoder (MAE) self-supervised pre-training procedure and find that a pre-trained USat outperforms state-of-the-art self-supervised MAE models trained on remote sensing data on multiple remote sensing benchmark datasets (up to 8\%) and leads to improvements in low data regimes (up to 7\%).
% Code and pre-trained weights are freely available at \url{link_redacted_for_anonymous_submission}.
Code and pre-trained weights are available at \url{https://github.com/stanfordmlgroup/USat}.
\end{abstract}

%% file: sec/1_introduction.tex
\section{Introduction}

The development of AI techniques for interpreting remotely sensed imagery has the potential to enable large-scale solutions for many domains with significant societal impact, including agriculture \cite{jung2021potential,nakalembe2023considerations}, energy \cite{ren2022automated}, climate and weather prediction \cite{bochenek2022machine}, disaster response \cite{kuglitsch2022facilitating}, and sustainable development \cite{burke2021using}.
% As remotely sensed imaging technology and AI techniques continue to improve, many more impactful applications will become possible. 
The biggest bottleneck to building and deploying these technologies is their dependency on labeled data, which substantially prohibits the development of such technologies for new regions and new tasks \cite{zhu2017deep,ma2019deep}. 
% This exacerbates bias towards regions with lots of available data and away from developing areas and communities which could benefit from these solutions the most.

\begin{figure*}[t!]
    \centering
    \includegraphics[width=\textwidth]{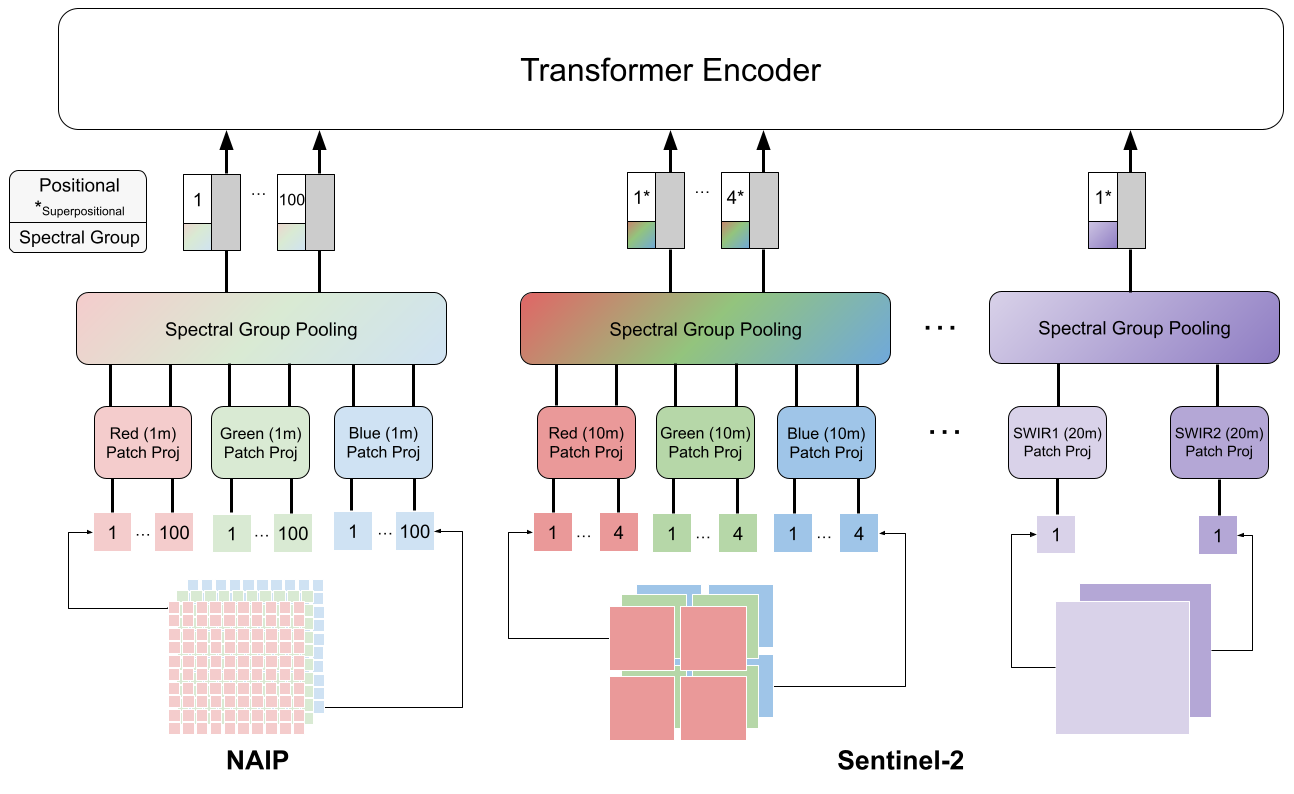}
    \caption{Overview of the USat encoder architecture. USat can accept any subset of spectral bands (channels) and spatial patches from multiple satellite (Sentinel-2) and aerial imagery (NAIP) sensors. Each spectral band is independently patchified, with lower GSD (higher spatial resolution) bands divided into more patches than higher GSD (lower spatial resolution) bands. We embed each band with a separate patch projection layer whose outputs are then input to the spectral group pooling layer which combine corresponding patches from different bands to produce the per-patch embeddings. Each patch embedding is then summed with an encoding vector which captures the positional and spectral group information of each patch, where superpositional encodings are used for the higher GSD bands indicated with *, before being fed into a Transformer. Best viewed in color.}
    \label{fig:usat}
\end{figure*}

The recent success of natural image and text foundation models in substantially reducing the need for labeled data presents a clear opportunity to adapt similar techniques to remotely sensed imagery \cite{lacoste2021toward,mai2023opportunities}. Foundation models are commonly pre-trained in a self-supervised fashion with large amounts of unlabeled data and subsequently fine-tuned for new tasks with limited labeled data \cite{bommasani2021opportunities}. The rich structure of remotely sensed imagery with multi-sensor, multi-spectral, and temporal data provides an immense amount of information that can be leveraged by self-supervised learning. Furthermore, this data is rapidly growing, with petabytes of new data generated each day and only increasing due to new satellites with higher coverage and finer spatial, spectral, and temporal resolutions \cite{reichstein2019deep}. 

Several prior works have begun leveraging the incredible amounts of remotely sensed imagery to build self-supervised approaches. Many of these studies developed contrastive learning approaches for remote sensing data \cite{wang2022self}, where positive pairs are constructed using vanilla image augmentations \cite{tao2020remote,li2022global}, geographic proximity \cite{jean2019tile2vec,kang2020deep,jung2021contrastive}, views from multiple sensors \cite{jain2021multi,swope2021representation,jain2022multimodal,chen2021self}, temporal imagery capturing the same geo-location \cite{manas2021seasonal,ayush2021geography}, and contrasting geo-tagged images with their geo-locations \cite{mai2023csp}. These methods have been shown to be effective for many different types of tasks and sensors, but depend on defining positive pairs which is often unclear how to do and can lead to harmful invariances in some cases \cite{xiao2020should}. Masked autoencoding (MAE) methods for remotely sensed imagery have emerged more recently, including a model to leverage temporal and multi-spectral imagery \cite{cong2022satmae}, an approach to incorporate scale-specific information in the model using an additional positional encoding \cite{reed2022scale}, a method which combines the MAE objective with a teacher-student distillation objective \cite{mendieta2023gfm}, and approaches designed for hyperspectral imagery \cite{ibanez2022masked, scheibenreif2023masked}. The vast majority of previous self-supervised approaches for remotely sensed imagery learn encoders that input images from a single sensor (e.g. a single satellite product)
% or only work with very specific sets of spectral bands,
thereby limiting their practical use and performance, a result we demonstrate in this work. 

Developing self-supervised methods that can leverage data from multiple sensors has the potential to provide additional self-supervision and allow for more tailored adaptation to downstream tasks. The success of contrastive learning methods which leverage multi-sensor views provides evidence that self-supervision from more than one sensor improves upon single sensor approaches \cite{jain2021multi,swope2021representation,jain2022multimodal,chen2021self}. SatMAE can naturally be extended to multiple sensors by including additional channel groups, but this results in increased sequence length leading to slower training and increased memory footprint, and in some cases unstable optimization \cite{cong2022satmae}. The key observation we make in this work is that different sensors can capture vastly different amounts of information due to differences in ground sampling distance (GSD). GSD captures the spatial scale of remotely sensed imagery using the linear distance between the centers of two adjacent pixels as measured on the Earth's surface. GSD can differ greatly between remotely sensed imaging sensors (e.g. from 0.3m Maxar imagery to 1km MODIS imagery) and even between spectral bands captured by the same sensor (e.g. Sentinel-2 has 10m, 20m, and 60m GSD bands). Most prior approaches directly rescale images before feeding them into the network which can unnecessarily increase model parameters and potentially hinder the ability of the model to leverage spatial scale which has been shown to benefit MAE models for remote sensing data in recent work \cite{reed2022scale}.
To solve these challenges, our primary contributions are:
\begin{enumerate}
    \item We design a new encoder, which we call USat, that can handle an arbitrary collection of images from multiple imaging products, each with different sets of spectral bands and ground sampling distances. Importantly, the encoder follows a simple design which substantially reduces sequence length, and consequently memory footprint and runtime, while preserving the geospatial alignment of images from different sensors.
    We integrate USat into a self-supervised MAE procedure called USatMAE to leverage the strong self-supervision signal present in multiple sensors.
    % USatMAE is a generalization of SatMAE, a previously developed state-of-the-art pre-training approach for satellite imagery.
    \item We introduce USatlas, a new benchmark derived from Satlas for developing and evaluating multi-sensor pre-training approaches, and conduct a wide variety of experiments testing the impact of using multiple sensors as well as different positional encodings on performance. We find that using multiple sensors during pre-training outperforms variants trained only with the low resolution sensor on all tested datasets, and that the encodings are key to further improved performance.
    \item Finally, we show that USatMAE outperforms all tested single sensor pre-training approaches on three of the four downstream remote sensing benchmarks and performs close to the best on the fourth.
    % \item We additionally investigate the use of different downstream task fine-tuning settings including different sensors and subsets of spectral bands, image size, and generalizing to new sensors and bands not present in the pre-training data. These experiments provide insights about how to fine-tune USat pre-trained models on new datasets.
\end{enumerate}
All code and pre-trained weights are freely available at\\ {\small \url{https://github.com/stanfordmlgroup/USat}}.

%% file: sec/2_methods.tex
\section{Methods}
In this section, we describe the USat encoder architecture and the USatMAE pretraining architecture.

\subsection{USat Encoder}
\label{usat}

The USat encoder uses a vision transformer (ViT) \cite{dosovitskiy2020image} with modified patch embeddings (Section~\ref{sec:patch}) and encodings (Section~\ref{sec:encodings}) to capture information in spectral bands of varying GSD across multiple sensors (Figure~\ref{fig:usat}).

\subsubsection{Patch Embeddings and Spectral Group Pooling}
\label{sec:patch}
A vanilla ViT uses a single patch embedding layer for the RGB channels together, but remotely sensed images commonly have many spectral bands (channels) each capturing different types of information at varying GSDs. Prior work has handled this by grouping bands of the same GSD together and using a different patch projection layer per group \citet{cong2022satmae}, but this design requires that all spectral bands within a group are present at fine-tune-time. This limits usability as practitioners may not have access to some bands or may not have computational resources to process and input all of them, which is partially why it is common to construct `false color composites' that are composed of three other bands used in place of the RGB bands.

To enable more input band flexibility, instead of using a patch projection for each group, we use a separate patch projection layer for each spectral band which allows the encoder to input any subset of spectral bands. As noted in \cite{cong2022satmae}, doing this naively leads to very long sequences, which substantially increases memory usage and can lead to unstable optimization. For example, if using 16x16 patches for each of the 13 Sentinel-2 bands, the sequence length is 3,328. However, using the same number of patches for bands of varying GSD wastes computation and model capacity.

Based on these observations, we design USat with two key modifications. First, we use a larger number of patches for lower GSD (higher spatial resolution) bands and a lower number of patches for higher GSD (lower spatial resolution). The number of patches per band are hyperparameters, with the constraint that highest GSD patch numbers must be divisble by the lower GSD patch numbers for the positional encodings (see the Encodings section below). With this modification, more tokens are used to capture the detailed information in higher GSD spectral bands while conserving memory on the lower GSD bands. However, adopting this modification alone still results in long sequence lengths which leads to runtime and memory footprint bottlenecks, motivating the next modification.

Second, we utilize spectral band grouping as done in prior work \cite{cong2022satmae}, but rather than enforcing that all bands are present before applying the patch embedding layer, we compute patch embeddings from individual bands separately and then aggregate after using a pooling over the patches in the same spectral group, which we refer to as `Spectral Group Pooling.' We note that spectral bands often have redundant information with other bands, so pooling can help reduce capacity without substantial reduction in expressiveness. We experiment with both sum pooling and average pooling (see Table~\ref{tab:pooling} in the Supplementary Material) and choose to proceed with average pooling as this naturally handles the inputting of arbitrary subsets of spectral bands without substantially affecting the scale of the output. It is worth noting that sum pooling is functionally equivalent to performing a spectral grouped patch embedding, and that average pooling is equally as expressive but is not equivalent due to weight regularization. The USat patchification, patch projection layers, and spectral group pooling layers are visualized in Figure~\ref{fig:usat}.

\subsubsection{Encodings}
\label{sec:encodings}
As the Transformer layers do not natively retain any positional information, we explore the use of positional encodings $E_P$ to provide the model information about the location of each patch, spectral group encodings $E_{SP}$ to provide spectral group information, and sensor encodings $E_{SE}$ to provide sensor information, each described below.

\noindent \textit{Positional Encodings} \,
We adopt the sine-cosine positional encodings as originally used in \cite{dosovitskiy2020image} to encode the 2D position of each patch:

\begin{align}
E_P(pos, 2i) = \sin(\frac{pos}{\Omega^{\frac{2i}{d}}}),\text{ }E_P(pos, 2i+1) = \cos(\frac{pos}{\Omega^{\frac{2i}{d}}})
\label{eq:pos}
\end{align}

\noindent where $pos$ is the position of the patch, $i$ is the index of the encoding, $\Omega$ is a large constant (usually set to 10000), and $d$ is the encoding dimension. However, by using independent patch embeddings, patches no longer represent the same area. For example, if the image captures a land cover of 320m x 320m, a spectral band A with 1m GSD and patch size of 16 x 16 has patches with a ground cover of 20m x 20m, while a spectral band B with 10m GSD and patch size of 8 x 8 has patches with a ground cover of 40m x 40m. The positional encodings should carry explicit information that each patch in spectral band B is positionally related to the four patches in spectral band A.

To do this, we propose the use of superpositional encodings. For the highest GSD spectral bands, we compute positional encodings by taking the average of the positional encodings from the lowest GSD patches that lie within the area captured by each patch. This results in the constraint that the number of patches in the higher GSD bands must divide the number of patches in the lowest GSD bands. For the spectral bands with the highest number of patches (the ones with the lowest GSD), we compute positional encodings following Equation \ref{eq:pos}. For the lowest GSD spectral groups, we use vanilla positional encodings.
% We hypothesize that the superpositional encodings allow the encoder to leverage the positional information of patches with varying ground cover.
See Figure~\ref{fig:superpositional_viz} in the Supplementary Material for a visualization of the superpositional encoding computation and Figure~\ref{fig:superpositional_sim} for an example visualization of the similarities between superpositional encodings and positional encodings.

\noindent \textit{Spectral Group and Sensor Encodings} \, 
In addition to positional encodings, we explore whether using spectral group encodings (as used in \cite{cong2022satmae}) $E_{SP}$ and sensor encodings $E_{SE}$ improve performance of the models. Specifically we use:

\begin{align}
E_{SP}(sp, 2i) = \sin(\frac{sp}{\Omega^{\frac{2i}{d}}}) \quad E_{SP}(sp, 2i+1) = \cos(\frac{sp}{\Omega^{\frac{2i}{d}}})
\end{align}
$sp$ is the index of all unique spectral groups, $i$ is the index of the encoding, $\Omega$ is a large constant (usually set to 10000), and $d$ is the encoding dimension. The formula for $E_{SE}(s, 2i)$ is the same except $s$ is the index of all unique sensors. These encodings provide information to the model about which spectral group (or which sensor) each patch corresponds to. 
% However, we note that the superpositional encodings are technically sufficient for the model to distinguish spectral groups with differing GSDs.
The total encoding size is 1024 for all experiments, and when using multiple encodings, we concatenate them for a total size of 1024, where 768 of the elements are positional and the remaining elements are allocated to either spectral group or sensor encodings, or half of each when using both. When using spectral group encodings, the spectral group encodings from the same group are different between sensors (i.e. even if the bands are the same between sensors, they use different spectral group encodings). 

\subsection{USatMAE}
\label{usatmae}

\subsubsection{USatMAE Decoder}
The USatMAE decoder inputs the USat encoder representations of unmasked patches along with mask tokens representing masked patches, and outputs a reconstruction of the masked patches, similar to the original MAE approach \cite{he2022masked}. Our decoder also uses an 8-layer transformer architecture and mean squared error reconstruction loss. The only difference is the positional encodings described in Section~\ref{sec:encodings}.

\subsubsection{Masking Strategy}
For our core experiments, we apply a random spatial mask to each band independently, which was shown in prior work (referred to as inconsistent masking) to outperform applying the mask identically to each band \cite{cong2022satmae}. However, in our work the number of patches and sizes of patches differs between each spectral group, so we account for this by ensuring equal ground cover areas are masked between bands. Specifically, we pre-compute a per-band mask number using the masking ratio and number of patches: $\lfloor p^2 \cdot r\rfloor$ which differs from band to band depending on the GSD of the band and allows for the same amount of ground cover to be masked across all products and bands.

\subsection{Pre-Training Dataset}
% We use BigEarthNet \cite{sumbul2019bigearthnet}, Satlas \cite{bastani2022satlas}, and METER-ML \cite{zhu2022meter} for pre-training and downstream evaluation, chosen based on the presence of multiple sensors in each dataset and consistency of GSD and ground cover between sensors.

We use the Satlas dataset for pre-training USatMAE \cite{bastani2022satlas}. Satlas consists of NAIP and Sentinel-2 imagery labeled for many different tasks. The NAIP images have 3 spectral bands (Red, Green, Blue) each either natively 0.6m or 1m GSD, and the Sentinel-2 images have 4 spectral bands of native 10m GSD (Red, Green, Blue, NIR) and 5 spectral bands of native 20m GSD (Red Edge 1-3, SWIR-2) upsampled with billinear interpolation to 10m GSD. The imagery is unpaired in its original form, so we pair the imagery and preprocess it using the following procedure:

\begin{enumerate}
    \item We match each NAIP image  with all the Sentinel-2 images that capture the same ground area and select the pair with the smallest absolute time difference.
    \item We crop the Sentinel-2 images to the same ground area as the paired NAIP image.
    \item We select the pairs that have the 3 NAIP bands and 9 Sentinel-2 bands.
    \item We resize all bands to their native 1m, 10m, or 20m GSD using bilinear downsampling. 
\end{enumerate}

To enable evaluation of pre-trained models on Satlas without task heads customized to specific backbone architectures, we convert Satlas to a multi-label classification task. Specifically, if any of the annotations (e.g. image-level label, point, polyline, polygon) corresponding to a class are present in the image, the image is labeled positive for the class, otherwise it is labeled negative. This results in 58 labeled tasks in the dataset.

We follow the same training, validation, and test splits as \citet{bastani2022satlas}. The final dataset consists of is 3,615,184 image pairs for training and 231,626 pairs for validation. In the validation set, we note that only 44 of the classes have at least one image labeled as positive for the class (see Figure~\ref{fig:usatlas} in the Supplementary Material for class counts in the training and validation sets). We evaluate performance on the dataset using micro-average precision (mAP) across the 44 classes on the validation set. To help facilitate the development of multi-sensor pre-training methods, we release the processing code to generate this modified version of Satlas, which we refer to as \textbf{USatlas}. 

% \textbf{Functional Map of the World.} \, The Functional Map of the World (fMoW) consists of 363,572 sub-meter resolution images labeled for the presence of 62 land cover and land use classes (cite). We pair this dataset with the fMoW-Sentinel dataset constructed in cite which adds all spectral bands of Sentinel-2 imagery to every example. We maintain all images with GSD <= 0.8m resulting in 358,296 pairs of images in total spanning more than 200 countries. We crop the images to their single bounding box annotation and assign the bounding box label to the image. 

\begin{table}[t!]
    \centering
    \begin{tabular}{c|c|c|c}
    \hline
        Superpositional & Spectral Group & Sensor & mAP \\
        \hline
        - & - & - & 65.08\\
        - & - & \checkmark & 65.77 \\
        - & \checkmark &  - & 66.88 \\
        - & \checkmark &  \checkmark & 65.81\\
        \checkmark & - & - & 66.33\\\
        \checkmark & - & \checkmark & 66.00 \\
        \checkmark & \checkmark & - & \textbf{66.99}\\
        \checkmark & \checkmark & \checkmark & 66.46  \\
        \hline
    \end{tabular}
    \caption{USatlas validation set performance when using different combinations of superpositional, spectral group, and sensor encodings. We pre-train with both sensors and fine-tune with Sentinel-2 for these experiments.}
    \label{tab:positional_ab}
\end{table}

\subsection{Downstream Datasets}
We use common remote sensing benchmarks BigEarthNet \cite{sumbul2019bigearthnet} and EuroSAT \cite{helber2019eurosat} for downstream evaluation, as well as METER-ML \cite{zhu2022meter}, all chosen based on the presence of Sentinel-2 and/or NAIP imagery in each dataset with a consistent ground cover captured by each image. We note that Sentinel-2 has global coverage with short revisit time (every 2-10 days) whereas NAIP only has coverage in the contiguous U.S., so Sentinel-2 is much more widely used. The dataset statistics, processing, and corresponding evaluation protocols are described in the Supplementary Material.

% \subsection{Transfer Learning Datasets}
% We use NAIP and the EuroSAT \cite{helber2019eurosat} datasets to evaluate the transfer learning performance of the models, selected based on use by prior works and the presence of sensors used for pretraining.

% \textbf{NAIP} \, NAIP contains TODO RGB images captured by the USDA’s National Agricultural Imagery Program and labeled for the presence of TODO land cover classes. We follow the same training, validation, and test splits as TODO. We evaluate performance on NAIP using accuracy.

% With yes, no, no on NAIP, pretrain both scores 71.71

%% file: sec/3_experiments.tex
\section{Experiments and Results}
We run several experiments testing the effect of different aspects of USatMAE and compare USatMAE to several baselines. All experiments use a ViT-L architecture for comparability. The pre-training and fine-tuning procedures are described in the Supplementary Material.

\subsection{Impact of superpositional, spectral group, and sensor encodings}
We investigate the impact of the superpositional, spectral group, and sensor encodings on USatlas validation set performance. Specifically, we pre-train SatMAE with every combination of encoding settings and fine-tune with the corresponding setting on the USatlas classification task. Due to computational constraints, for these experiments we use 10\% of the USatlas training set for pre-training and fine-tuning, and fine-tune with Sentinel-2 alone as Sentinel-2 is directly affected by the superpositional, spectral group, and sensor encoding variations.

We find that the use of superpositional encodings outperforms the use of regular positional encodings across all comparable encoding settings by an average of +0.56 mAP (Table~\ref{tab:positional_ab}). The performance gap is largest (+1.25 mAP) when using no spectral group and no sensor encodings, which suggests the model may be learning to leverage the positional encodings from different GSDs to identify the spectral group or sensor. Furthermore, the use of spectral group encodings leads to performance improvements across all combinations of encodings (+0.74 on average).  In general, using a sensor encoding decreases performance (by an average of -0.31 mAP) excluding when using no superpositional encodings and no spectral group encodings.  We find that this improvement does not hold when using NAIP for fine-tuning, likely because NAIP has one spectral group whereas Sentinel-2 has multiple (see Table~\ref{tab:sensor_encoding_ab} in the Supplementary Material). Finally, when fine-tuning with Sentinel-2 alone, we surprisingly find that using the spectral group encodings corresponding to the new indexing used during fine-tuning, rather than the original indexing used during pre-training, leads to consistent small improvements across all tested settings and datasets (see Section~\ref{sec:group_ab} in the Supplementary Material for further discussion). We use the best encoding setting based on Sentinel-2 (superpositional, spectral group, no sensor) for all subsequent experiments.

\begin{table}[!t]
    \centering
    \begin{tabular}{c|ccc}
    \hline
    \multirow{2}{*}{Pre-training Sensor} & \multicolumn{3}{c}{Downstream Sensor}\\
    & NAIP & Sentinel-2 & Both\\
    \hline
    % No Pre-training & 18.17 & 15.35 & 18.30\\
    % NAIP & 20.24 & - & -\\
    % S2 & - & 14.12 & -\\
    % Both & \textbf{20.81} & \textbf{15.54} & \textbf{21.37}\\
    No Pre-training & 67.73 & 65.32 & 66.67\\
    NAIP & \textbf{71.21} & - & -\\
    Sentinel-2 & - & 66.50 & -\\
    Both & 70.81 & \textbf{67.62} & \textbf{72.53}\\
    \hline
    \end{tabular}
    \caption{Comparison between multi-sensor and single sensor pre-training on USatlas. We report mAP across the USatlas validation set tasks.}
    \label{tab:sensor_transfer}
\end{table}

\begin{table*}[t!]
    \centering
    \begin{tabular}{c|c|cccc}
    \hline
    % \multirow{2}{*}{Initialization} & \multicolumn{3}{c|}{fMoW} & \multicolumn{3}{c|}{BigEarthNet} & \multicolumn{3}{c|}{Satlas} & \multicolumn{4}{c}{METER-ML}\\
    \multirow{2}{*}{Pre-Train Strategy} & \multirow{2}{*}{Pre-Train Dataset} & EuroSAT (Acc) & BigEarthNet (mAP) & \multicolumn{2}{c}{METER-ML (MAP)}\\
    & & Sentinel-2 & Sentinel-2 & Sentinel-2 & NAIP\\ 
     % & Maxar & S2 & Both & S2 & S1 & Both & NAIP & S2 & Both & NAIP & S2 & S1 & All\\
    \hline
    % Random Init & - & 74.71 & 94.81 & 35.11 & 51.37\\
    Random Init & - & 96.70 & 78.33 & 47.44 & 69.41\\
    % Supervised & ImageNet & 74.50 & 94.96 & 34.67 & 82.81\\
    % MAE \cite{he2022masked} & ImageNet & 77.95 & 96.44 & 44.09 & 79.62\\
    % SatMAE \cite{cong2022satmae} & fMoW & 80.61 & 98.06 & 49.87 & 71.83\\
    SatMAE \cite{cong2022satmae} & fMoW Sentinel & 97.65 & 84.40 & \textbf{68.46} & 76.87\\
    % CSF & ViT-L &  & -\\
    % ScaleMAE \cite{reed2022scale} & fMoW & 78.87 & 96.09 & 49.71 & 82.12\\
    ScaleMAE \cite{reed2022scale} & fMoW RGB & \textbf{98.59} & 84.24 & 68.42 & 78.38\\
    % SatlasNet & Swin-Base & pull & -\\
    \hline
    % USatMAE (Ours) & USatlas S2 & 81.09 & 95.98 & 38.32 & -\\
    USatMAE (Ours) & USatlas Sentinel-2 & 97.06 & \textbf{85.09} & 66.59 & -\\
    % USatMAE (Ours) & USatlas NAIP & - & - & - & 80.03\\
    USatMAE (Ours) & USatlas NAIP & - & - & - & \textbf{83.68}\\
    % USatMAE (Ours) & USatlas Both & \textbf{81.30} & \textbf{98.39} & \textbf{55.69}& 81.21\\
    USatMAE (Ours) & USatlas Sentinel-2 + NAIP & \textbf{98.37} & \textbf{85.82} & \textbf{73.95} & \textbf{83.50}\\
    \hline
    \end{tabular}
    \caption{Performance of USatMAE compared to baselines on the test set of each downstream task. All approaches are trained using a ViT-L architecture. We evaluate on EuroSAT using accuracy (Acc), on BigEarthNet using micro-average precision (mAP), and on METER-ML using macro-average precision (MAP). We bold the top two performing approaches in each column.}
    \label{tab:baselines}
\end{table*}

\subsection{Comparison between multi-sensor and single sensor training}
We investigate the benefit of multi-sensor pretraining and fine-tuning compared to using single sensor on USatlas. Specifically, we compare single sensor pre-training (NAIP alone, Sentinel- alone) against multi-sensor pre-training (both NAIP and Sentinel-2) and no pre-training. We evaluate the single sensor pre-training runs by fine-tuning with the same sensor and evaluate the multi-sensor pre-training and no pre-training by fine-tuning with each sensor individually and together as separate experiments. For these experiments we use all of the USatlas training dataset for pre-training and 10\% for fine-tuning.

USatMAE pre-trained with multiple sensors outperforms the single sensor variants and no pre-training on USatlas when using all combinations of downstream sensors (Table~\ref{tab:sensor_transfer}). When fine-tuning with Sentinel-2 alone, the multi-sensor pre-training achieves +1.12 mAP higher than pre-training with Sentinel-2 alone which suggests that the joint pre-training with the higher resolution sensor (NAIP) improves representations of the lower resolution sensor (Sentinel-2). It also leads to +2.30 mAP higher than no pre-training. Qualitatively, the multi-sensor pre-trained model Sentinel-2 reconstructions show much higher quality and fidelity to the original image compared to the Sentinel-2 pre-trained model (see Figure~\ref{fig:inpainting_s2} in the Supplementary Material). 

When fine-tuning with NAIP alone, the multi-sensor pre-training achieves -0.4 mAP lower than pre-training with NAIP which suggests that the added spectral information from Sentinel-2 may not benefit the NAIP representations (also supported by the similar quality NAIP reconstructions between the two models shown in Figure~\ref{fig:inpainting_naip}).
% However, this performance difference is small relative to the other differences. 
Multi-sensor pre-training leads to a +3.08 higher mAP compared to no pre-training when fine-tuning on NAIP. Pre-training with both sensors and fine-tuning with both sensors achieves the highest overall performance (72.53 mAP), outperforming the NAIP fine-tuning by +1.32 mAP and both fine-tuning with no pre-training by +5.86 mAP. Finally, to demonstrate the flexibility of USatMAE in working with arbitrary spectral bands, we experiment with using different subsets of spectral bands when fine-tuning the multi-sensor pre-trained model on USatlas with Sentinel-2 alone (see Table~\ref{tab:spectral_bands} in the Supplementary Material). 

% \subsubsection{Impact of sensor and spectral encodings}

% \subsubsection{Impact of pretraining data}
% Table 3
% Usat enables pre-training on a combination of datasets with different sensors. Could do this with SatMAE / ScaleMAE too? How to make a fair comparison?

% Also can we do some scale tests here?

% \begin{table}[t!]
%     \centering
%     \begin{tabular}{c|ccc}
%     \hline
%     \multirow{2}{*}{Pretraining Data} & \multicolumn{3}{c}{Fine-tuning Data}\\
%     % & fMoW & BigEarthNet & Satlas & METER-ML\\
%     & BigEarthNet (mAP) & Satlas (IoU) & METER-ML (mAP)\\
%     \hline
%     % fMoW &  \\
%     BigEarthNet & \\
%     Satlas & \\
%     METER-ML & \\
%     All & \\
%     \hline
%     \end{tabular}
%     \caption{Impact of pretraining data on the validation set performance of each task. All experiments use all available sensors at pretrain-time and fine-tune-time.}
%     \label{tab:dataset_transfer}
% \end{table}

% \subsection{Effect of model and dataset size}
% We measure the effect of model and dataset size on the performance of USatMAE. Specifically, we experiment with ViT-S, ViT-B, and ViT-L architectures each using varying fractions of data in USatlas, incuding 1\% (~36k examples), 10\% (~360k examples), and 100\% (3.6M examples). All models are pre-trained and fine-tuned with both the NAIP and Sentinel-2 sensors.

% Dataset and model size substantially impact model performance (Figure~\ref{fig:data_model_scale}). TALK ABOUT TRENDS.

\begin{figure*}[t!]
    \centering
  \begin{subfigure}{.32\linewidth}
    \centering\includegraphics[width=1\linewidth]{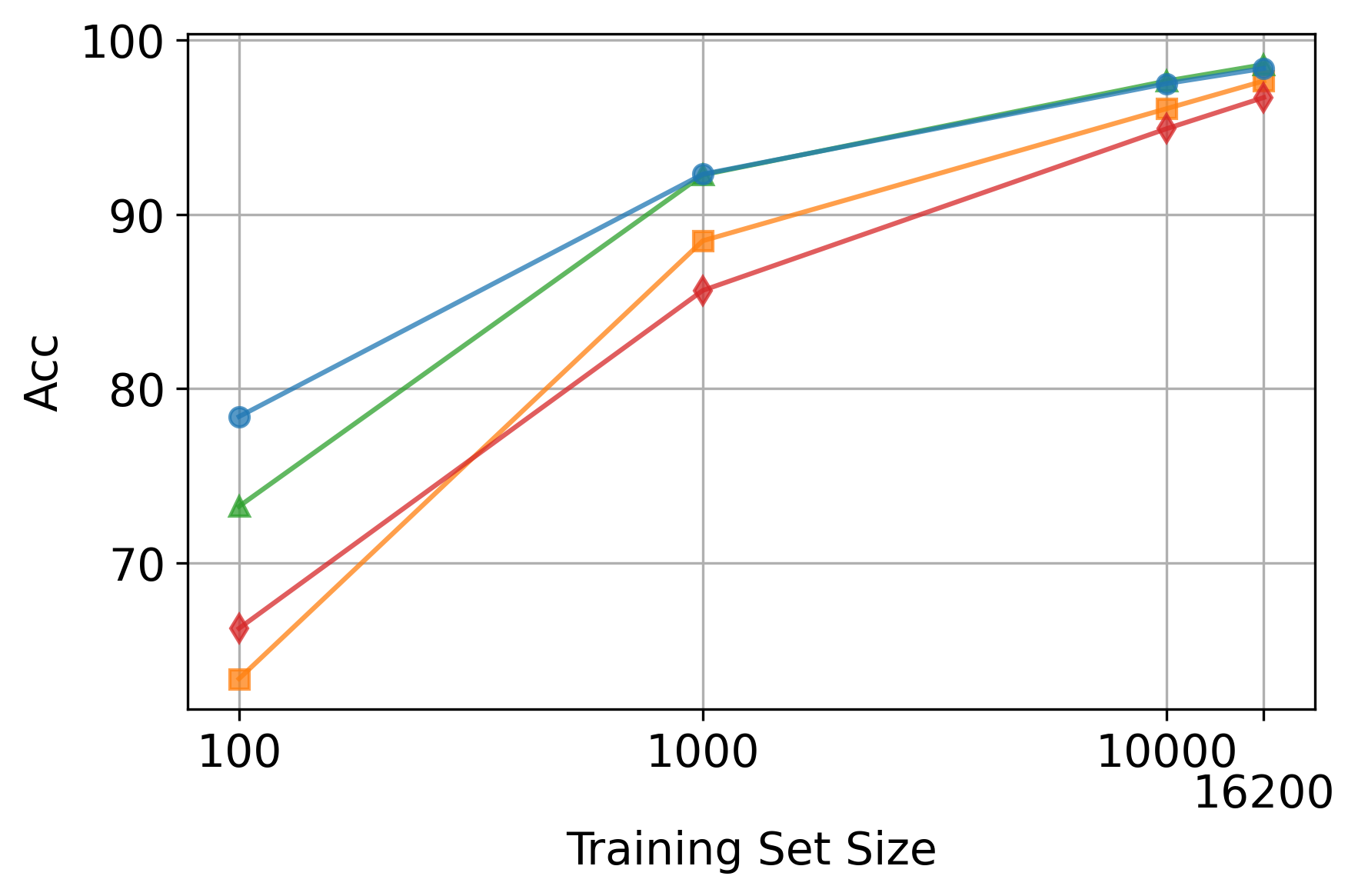}
    \caption{EuroSAT}
    \label{fig:num_examples_eurosat}
  \end{subfigure}
  \begin{subfigure}{.32\linewidth}
    \centering\includegraphics[width=1\linewidth]{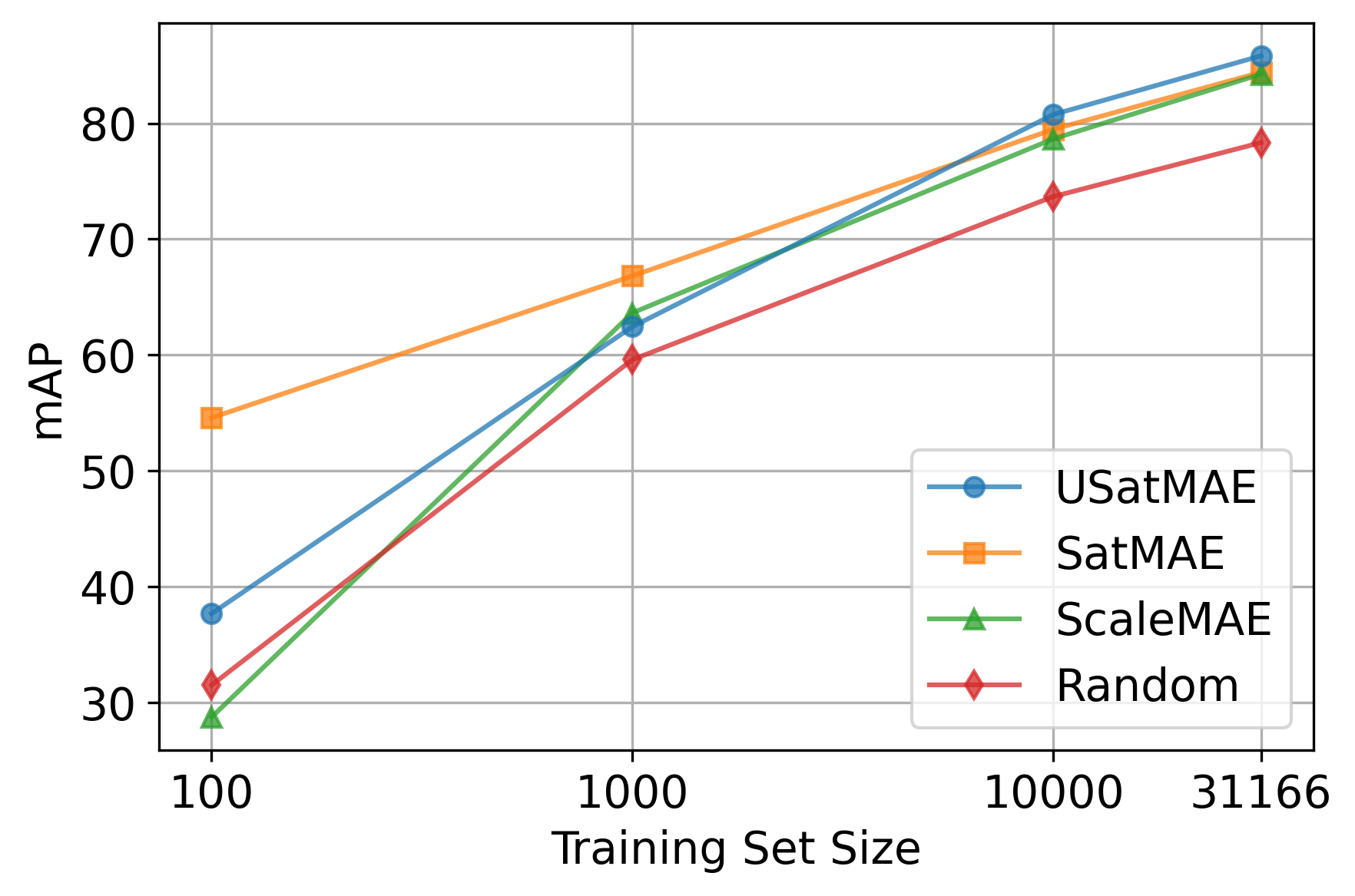}
    \caption{BigEarthNet}
    \label{fig:num_examples_ben}
  \end{subfigure}
  \begin{subfigure}{.32\linewidth}
    \centering\includegraphics[width=1\linewidth]{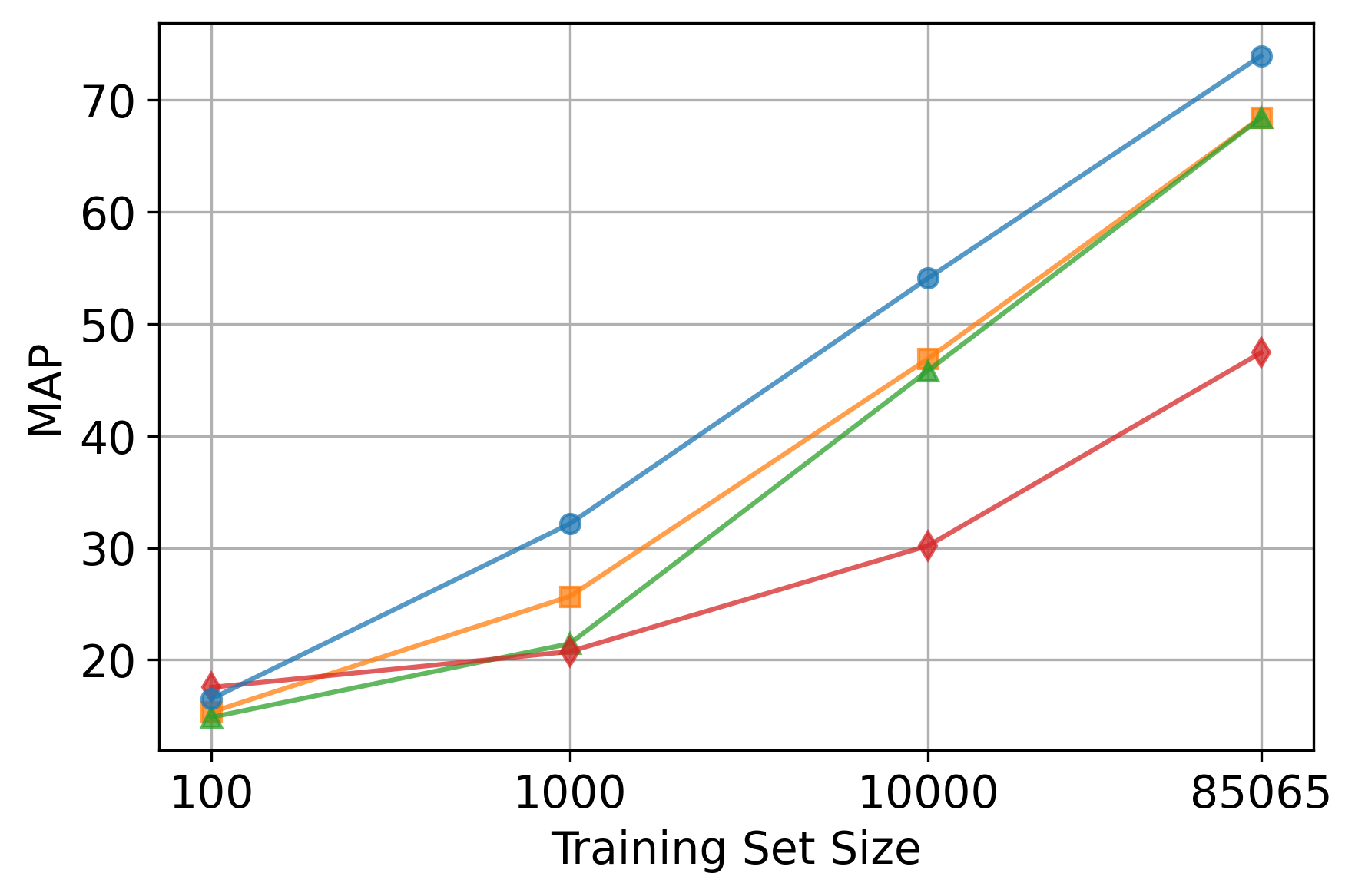}
    \caption{METER-ML Sentinel-2}
    \label{fig:num_examples_meterml}
  \end{subfigure}
    \caption{Test set performance of USatMAE against baselines models across varying training set sizes of downstream datasets. EuroSAT is evaluated using accuracy (Acc), BigEarthNet micro-average precision (mAP), and METER-ML macro-average precision (MAP).}
    \label{fig:num_examples}
\end{figure*}

\subsection{Comparison to baselines}
\subsubsection{Performance on the full downstream datasets}
We compare ViT-L USatMAE pre-trained on all of USatlas with both sensors using the best encoding setting against multiple self-supervised approaches which use remotely sensed imagery for pre-training including SatMAE \cite{cong2022satmae} and ScaleMAE \cite{reed2022scale} as well as a randomly initialized ViT-L. For these experiments, we fine-tune each method on the full downstream datasets. We use the public ViT-L checkpoints for both approaches. To fairly compare to ScaleMAE, we load the pre-trained RGB weights and randomly initialize the weights from the other bands when using multiple spectral bands in the Sentinel-2 experiments so that all methods use all input bands when fine-tuning.
% , and our re-implementation of contrastive sensor fusion (CSF, \citet{swope2021representation}) trained on USatlas.
We note we do not compare to SatlasNet \cite{bastani2022satlas} as it uses a Swin Transformer \cite{liu2021swin}, not a ViT-L. We additionally evaluate USatMAE pre-trained with each sensor individually to measure the benefit from using both sensors on the downstream datasets.

USatMAE trained with both sensors attains the best or second best result amongst all baseline models on the tested remote sensing benchmarks (Table~\ref{tab:baselines}). On EuroSAT, it outperforms random initialization by +1.67\% and SatMAE by +0.72\% accuracy, but underperforms ScaleMAE by -0.22\% accuracy. On BigEarthNet, USatMAE trained with both sensors outperforms all baseline methods, specifically random initialization by +7.49 mAP, SatMAE by +1.42 mAP, and ScaleMAE by +1.78 mAP. USatMAE trained with both sensors on METER-ML achieves the highest performance among the baseline models both when fine-tuning with Sentinel-2 and with NAIP. Specifically, USatMAE outperforms SatMAE (+5.49 MAP for Sentinel-2, +6.63 MAP for NAIP) and ScaleMAE (+5.53 MAP for Sentinel-2, +5.12 for NAIP), and substantially outperforms random initialization (+26.51 MAP for Sentinel-2, +14.09 MAP for NAIP). SatMAE achieves the second highest performance of the three baselines on BigEarthNet and METER-ML with Sentinel-2 imagery whereas ScaleMAE achieves the highest performance on EuroSAT and the second higher on METER-ML with NAIP. All methods outperform random initialization, with the largest performance differences on METER-ML followed by BigEarthNet.

USatMAE trained with Sentinel-2 underperforms USatMAE trained with both sensors by -1.31\% accuracy on EuroSAT and -7.36 MAP on METER-ML Sentinel-2, and also underperforms both SatMAE and ScaleMAE, indicating the benefit from including both sensors during pre-training, and that it is not the use of the USatlas dataset alone that leads to the performance achieved by USatMAE on these datasets. On BigEarthNet, USatMAE trained with Sentinel-2 only also underperforms USatMAE trained with both sensors (-0.71 mAP) but SatMAE by +0.69 mAP, ScaleMAE by +0.84 mAP, and random initialization by +6.76 mAP. On METER-ML with NAIP imagery, single sensor pre-training slightly outperforms (+0.18 MAP) multi-sensor pre-training, consistent with the findings in Table~\ref{tab:sensor_transfer}.

\subsubsection{Performance at low data fractions}
We further compare performance against the baselines at varying numbers of examples in the Sentinel-2 downstream datasets. Specifically, we take a random sample of 100, 1,000, and 10,000 examples of each of the downstream training sets and then compare test set performance of models trained on these data subsets.

We find that USatMAE achieves the highest or second highest performance among all baselines across almost all tested training dataset sizes in the three datasets (Figure~\ref{fig:num_examples}). On EuroSAT, USatMAE outperforms all methods using both 100 examples (+5.15\%, +15.04\%, +12.15\% accuracy on ScaleMAE, SatMAE, and Random init respectively) and 1,000 examples (+0.04\%, +3.82\%, +6.67\% accuracy on ScaleMAE, SatMAE, and random init respectively). Performance differences between methods reduce substantially at larger training set sizes, on which USatMAE underperforms ScaleMAE by -0.15\% accuracy with 10,000 examples but outperforms SatMAE (+1.44\%) and random init (+2.59\%). 

On BigEarthNet, SatMAE achieves the highest performance using 100 and 1000 examples, outperforming USatMAE by +16.9 and +4.38 mAP respectively. USatMAE outperforms ScaleMAE (+8.94 mAP) using 100 examples and underperforms it (-1.19) using 1000 examples. USatMAE achieves the highest performance at the larger dataset sizes, including +1.28 and +2.1 mAP improvements over SatMAE and ScaleMAE respectively when using 10,000 examples. Random initialization underperforms all methods across all training set sizes except for ScaleMAE when using 100 examples. 

On METER-ML, performance differences are minimal between methods and close to random performance when using 100 examples likely due to insufficient data per class. USatMAE outperforms all methods by +10.7, +6.49, and +11.41 MAP using 1000 examples and by +8.24, +7.19, and +23.89 MAP using 10000 examples on ScaleMAE, SatMAE, and random init respectively.

\begin{table*}[!t]
    \centering
    \begin{tabular}{c|c|cccc}
        \hline
        \multirow{2}{*}{Pre-Train Strategy} & \multirow{2}{*}{Pre-Train Dataset} & EuroSAT (Acc) & BigEarthNet (mAP) & \multicolumn{2}{c}{METER-ML (MAP)}\\
        & & Sentinel-2 & Sentinel-2 & Sentinel-2 & NAIP\\ 
        \hline
        Supervised \cite{dosovitskiy2020image} & ImageNet & \textbf{99.15} & 84.86 & \textbf{77.37} & \textbf{80.53}\\
        MAE \cite{he2022masked} & ImageNet & \textbf{98.89} & \textbf{85.41} & 73.54 & 80.34\\
        \hline
        USatMAE (Ours) & USatlas Sentinel-2 + NAIP & 98.37  & \textbf{85.82} & \textbf{73.95} & \textbf{83.50}\\
        \hline
    \end{tabular}
    \caption{Comparison between USatMAE and ImageNet initialization on the test set of each downstream task. Both approaches are trained using a ViT-L architecture. We bold the top two performing models in each column.}
    \label{tab:imagenet}
\end{table*}

\subsubsection{Comparison of USatMAE to ImageNet Pre-Training}
The most common initialization for remote sensing tasks remains ImageNet pre-training, so we compare the performance of USatMAE to a ViT-L pre-trained on ImageNet in a fully supervised fashion \cite{dosovitskiy2020image} and self-supervised using masked auto encoding (MAE) \cite{he2022masked}. To fairly compare all approaches, we load the pre-trained RGB weights and randomly initialize the weights from the other bands when using multiple spectral bands in the Sentinel-2 experiments so that all input bands are used when fine-tuning.

Surprisingly, we find that both ImageNet pre-training methods are competitive with USatMAE across the benchmark datasets (Table~\ref{tab:imagenet}). Supervised ImageNet pre-training achieves the highest performance on EuroSAT (+0.78 Acc compared to USatMAE) and METER-ML with Sentinel-2 imagery (+3.42 MAP). MAE ImageNet pre-training underperforms USatMAE on everything but EuroSAT, where it outperforms USatMAE (+0.52 Acc). These findings differ from \cite{cong2022satmae} which we believe can likely be attributed to the use of the full set of spectral bands when fine-tuning, rather than just RGB, and non-standardized backbones. We note that these results are consistent with recent work which finds ImageNet pre-training to be competitive when using careful image sizing and normalization \cite{corley2023revisiting} and another work that finds self-supervised learning on ImageNet transfers well compared to in-domain pre-training \cite{calhoun2022self}. The strength of this approach and the ImageNet pre-trained MAE approach could also be attributed to the scale of pre-training data, which is more than 5x larger than the data used to pre-train USatMAE.  We note that USatMAE largely closes the gap between the MAE baselines and supervised ImageNet pre-training on METER-ML with Sentinel-2, although a nontrivial gap still remains. However, USatMAE outperforms both ImageNet pre-training methods on BigEarthNet (+0.96 mAP and +0.41 mAP compared to supervised and MAE respectively) and METER-ML (+2.97 MAP and +3.16 MAP compared to supervised and MAE respectively).

%% file: sec/4_discussion.tex
\section{Discussion}

\subsection{Related Work}
We highlight the differences from similar recent works here.

SatMAE also adapts a ViT to multi-spectral satellite imagery and pretrains it using a masked autoencoding objective \cite{cong2022satmae}. Our approach primarily differs by leveraging data from multiple sensors, using superpositional encodings, and treating different spectral bands independently, which allow the model to account for differing GSD and enable the use of any subset of spectral bands when fine-tuning. 

ScaleMAE similarly uses a ViT with MAE pretraining but includes a GSD positional encoding to leverage the spatial scale of the image and uses a modified decoder to reconstruct single sensor images with varying frequency at corresponding scales \cite{reed2022scale}. Our approach incorporates GSD information directly into the patchification and patch projection layers as well as superpositional encodings, and supports the use of multiple sensors with bands of varying GSD during pretraining and fine-tuning.

ClimaX customizes a ViT for weather and climate prediction and is pretrained using forecasting tasks \cite{nguyen2023climax}. It assumes a fixed resolution across all input products and handles long sequence lengths using a cross-attention channel aggregation scheme. USat is trained to reconstruct multi-spectral satellite and aerial imagery with spectral bands of varying GSD from a single time point.

Presto pretrains a Transformer to reconstruct multi-spectral remotely sensed imagery over time \cite{tseng2023lightweight}. Similar to USat, Presto can leverage any subset of multi-spectral inputs from multiple sensors. However, it differs from USat in several important ways: (1) Presto operates per-pixel, so it cannot natively input images with many spatial pixels.
% This means it does not leverage spatial structure during pre-training and cannot easily do so when fine-tuning, especially with images capturing large ground cover. For this reason, it is efficient and uses a small number of parameters, but for a 224x224 image for example, USat produces a result in a single forward pass but Presto would require more than 50,000 forward passes.
(2) Presto does not leverage any imagery with GSD higher than 10m, which means it may not work with higher resolution imagery and may not benefit from that rich self-supervision signal. (3) Presto uses both temporal and geographic location information, both of which can be integrated into USat,
% through positional encodings
but we leave this as a direction for future work.

\subsection{Limitations}
Our work has limitations. First, because USat can input multiple sensors jointly, it requires more memory and can be slower to train than single sensor pre-training methods. However, these constraints are reduced when doing single-sensor fine-tuning and it is possible that efficient fine-tuning methods can allow for much larger sequence lengths during fine-tuning \cite{chen2023longlora}. Still, this limitation makes it difficult to leverage temporal information during pre-training which has been shown to provide useful self-supervision signal in several prior works \cite{cong2022satmae,tseng2023lightweight,manas2021seasonal,ayush2021geography}. Second, in a multi-sensor setting, USat assumes a fixed ground area between sensors and divisible GSDs for the superpositional encodings but this is not always the case in remote sensing datasets. Third, we do not explore fine-tuning USatMAE for dense predictions tasks. It is possible to adapt USat to multi-scale model such as Swin \cite{liu2021swin} by replacing the patch embedding layer, but we leave this to future work.
Fourth, we only experiment with two sensors (NAIP and Sentinel-2), but there are many other sensors that can be used for pre-training with varying spatial, spectal, and temporal resolutions. Our approach is general enough to support an arbitrary amount of sensors but making this computationally feasible remains an open problem. Furthermore, testing USatMAE's generalizability to other sensors is an interesting future direction.
% Second, as USat preserves ground area information, it can also only accept a limited ground cover size due to memory constraints. We believe 720m x 720m is sufficient for many applications, and this can be increased if off-the-shelf if not using the lowest (1m) GSD imagery. Balancing this tradeoff between spatial, spectral, and temporal resolutions with ground cover size during both pre-training and fine-tuning is an interesting direction for future work.  

% \subsection{Societal Impacts}
% We believe our work has the potential to lead to substantial societal benefits by supporting the development of AI for remote sensing technologies for important domains like agriculture, disaster response, and climate change. However, we acknowledge potential negative impacts here.

% \subsubsection{Carbon Footprint} \, 

% \subsubsection{Bias} \, 

% \subsubsection{Privacy} \, 

%% file: sec/5_conclusion.tex
% \section{Conclusion}
% We develop a new architecture called USat and pretraining procedure called USatMAE designed for self-supervised learning with multi-spectral remotely sensed imagery from multiple sensors. Our proposed approach improves performance on several benchmark remote sensing datasets compared to other pre-training approaches. We hope that our work supports the development of remote sensing models for many AI for good applications. 
\section{Conclusion}
The USat encoder and USatMAE architecture are new approaches to leverage multi-sensor remotely sensed imagery in a flexible manner, allowing the model to benefit from complementary information from different sensors composed of spectral bands with varying ground sampling distance. Our results demonstate the benefit of the USatMAE pre-training approach on the USatlas dataset. Notably, we find USatMAE trained with both sensors outperforms the use of each sensor individually, showing the benefit of including both types of information. This has implications for developing models on Sentinel-2 imagery which is available globally but NAIP is not. Furthermore, USatMAE increases downstream flexibility by enabling fine-tuning with varying subsets of spectral bands. Finally, when jointly training using multiple sensors with different GSDs, USatMAE does not downsample the low GSD (high spatial resolution) as most prior works do, which preserves the image scale and avoids potential loss of valuable information in the images. We hope that our work supports the development of remote sensing models for many AI for good applications. 

%% file: sec/6_supplementary.tex
\clearpage
\setcounter{page}{1}
\maketitlesupplementary

% \section{Rationale}
% \label{sec:rationale}
% % 
% To split the supplementary pages from the main paper, you can use \href{https://support.apple.com/en-ca/guide/preview/prvw11793/mac#:~:text=Delete%20a%20page%20from%20a,or%20choose%20Edit%20%3E%20Delete).}{Preview (on macOS)}, \href{https://www.adobe.com/acrobat/how-to/delete-pages-from-pdf.html#:~:text=Choose%20%E2%80%9CTools%E2%80%9D%20%3E%20%E2%80%9COrganize,or%20pages%20from%20the%20file.}{Adobe Acrobat} (on all OSs), as well as \href{https://superuser.com/questions/517986/is-it-possible-to-delete-some-pages-of-a-pdf-document}{command line tools}.

\section{Downstream Datasets}

\subsubsection{EuroSAT} The EuroSAT dataset consists of 27,000 13-channel Sentinel-2 images labeled for the presence of 10 land cover classes \cite{helber2019eurosat}. Each image is of size 64 x 64 and covers a ground area of 640m x 640m.  We use the training, validation, and test splits defined by torchgeo \cite{stewart2022torchgeo} which have 16,200, 5,400, and 5,400 images respectively. We evaluate performance on EuroSAT using accuracy (Acc) following \citet{helber2019eurosat,cong2022satmae,reed2022scale}.

\subsubsection{BigEarthNet} BigEarthNet contains 590,326 paired Sentinel-2 and Sentinel-1 images spanning 10 countries and multi-labeled for the presence of 19 classes \cite{sumbul2019bigearthnet}. Each image is of size 120 x 120 covering a ground area of 1200m x 1200m. We apply equal padding of 4 pixels to all sides of each image, resulting in an image size of 128 x 128 to be compatible with the image sizes of the other datasets. All Sentinel-2 bands (except for B10, which was excluded from BigEarthNet) are included in the dataset. We filter out images with seasonal snow, cloud cover, and cloud shadow as recommended by the original authors \cite{sumbul2019bigearthnet}. We also discard the 60m GSD bands in our training runs per the findings from \cite{cong2022satmae}, leaving us with a total of 10 bands from Sentinel-2. We follow \citet{cong2022satmae,reed2022scale} and use the same training (a 10\% sample), validation, and test splits, and evaluate performance on BigEarthNet as a multi-label classification task using micro-average precision across the 19 classes on the test set. The resulting training, validation, and test sets have 31,166, 103,944, and 103,728 images respectively.

\subsubsection{METER-ML} METER-ML consists of 86,625 triplets of NAIP, Sentinel-2, and Sentinel-1 images multi-labeled for the presence of six infrastructure classes in the contiguous U.S. \cite{zhu2022meter}. METER-ML consists of 720 x 720 images capturing a 720m x 720m area on the ground. We use the same training, validation, and test splits as defined in \cite{zhu2022meter} which have 85,065, 515, 1,018 images respectively. We evaluate performance on METER-ML as a multi-label classification task using a macro-average precision (MAP) on the test set following \citet{zhu2022meter}.

\section{Pre-Training Procedure}
For all experiments, we establish a maximum image footprint of 1280m x 1280m in which all subsequent input images should fit as concentric squares. Based on the 320m x 320m USatlas footprint, we use 20x20 patches each with a patch size of 16x16 for the (1m) NAIP spectral bands, 4x4 patches each with a patch size of 8x8 for the 10m Sentinel-2 spectral bands, and 2x2 patches each with a patch size of 8x8 for the 20m Sentinel-2 spectral bands. We choose these patch sizes to be similar to sizes shown to be effective for general masked autoencoders \cite{he2022masked} and recent work applying them to satellite imagery \cite{cong2022satmae,reed2022scale}.

We use an AdamW optimizer \cite{loshchilov2017decoupled} with a cosine learning rate scheduler and base learning rate of 0.00015. We use a standardized batch size of 160 for Sentinel-2, NAIP, and both sensor pre-training to reduce batch size variability when comparing models. As our preliminary experiments suggested that performing inconsistent random spatial masking (where masking between spectral groups is different) was substantially better than other masking schemes like consistent random spatial masking and spectral group masking (consistent with findings in \cite{cong2022satmae}), we use inconsistent random spatial masking with a masking ratio of 0.75.  We use random horizontal and vertical flips. We do not use image re-scaling as an input augmentation to preserve the native GSD. We use the final checkpoint after 25 epochs. We train all models using 10 NVIDIA A4000 GPUs with the batch split across the GPUs using data parallelism.

\section{Fine-Tuning Procedure}
For datasets with image footprints smaller than the maximum image footprint of 1280m x 1280m (BigEarthNet, METER-ML, EuroSAT), we compute positional encodings as if the image is positioned in concentric squares inside of the maximum image footprint. This ensures that the positional encodings applied to the same geospatial location across different GSDs is the same. We fine-tune the network with consistent GSDs to pre-train time, although we note that it would be possible to fine-tune using a higher GSD at fine-tune-time than pre-train time by interpolating the positional encodings as is common practice for transferring to higher resolution natural images \cite{dosovitskiy2020image}. For USatlas, we use an AdamW optimizer with a cosine learning rate scheduler, a base learning rate of 0.001, and a weight decay of 0.1, and identical batch sizes to the pre-training setting. We train for 5 epochs but the model typically converges in the first 3 epochs, likely due to the large size of the dataset. For the downstream datasets, we use an AdamW optimizer with a cosine learning rate scheduler. We use a learning rate of 0.00001 for EuroSAT and METER and 0.0001 for BigEarthNet. We maximize batch size to fit on our hardware without using gradient accumulation. We use 10 epochs with 1 warmup epoch and a batch size of 40 for BigEarthNet, 50 epochs with 10 warmup epochs and a batch size of 50 for EuroSAT, and 5 epochs with 1 warmup epoch and a batch size of 10 for METER-ML. We select the best checkpoint over all epochs using the respective metric used for each dataset, macro-averaged over all validation set batches. We train all models using 10 NVIDIA A4000 GPUs with the batch split evenly across the GPUs using data parallelism.

\begin{figure}[!t]
    \centering
    \includegraphics[width=\columnwidth]{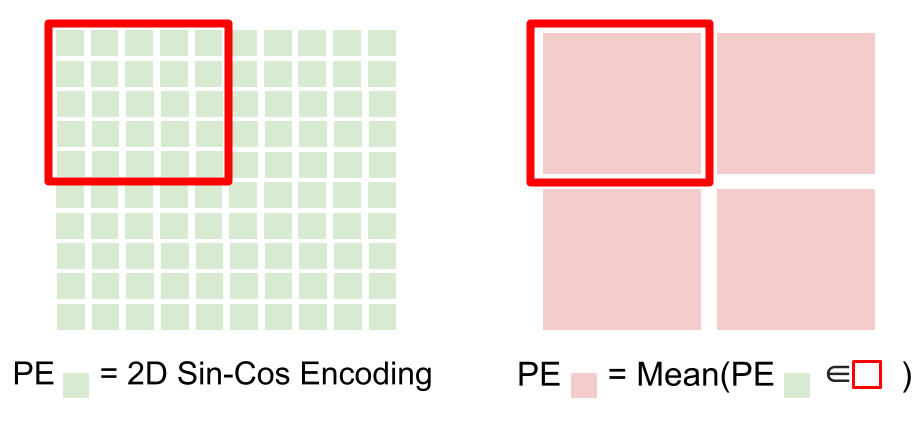}
    \caption{Example superpositional encodings used in USatMAE. Images from sensors capture identical ground areas but GSD can be different between sensors, so we use superpositional encodings to capture the positional relationships. Specifically for higher GSD (lower spatial resolution) sensors, we use the mean of the lower GSD (higher spatial resolution) positional encodings.}
    \label{fig:superpositional_viz}
\end{figure}

\subsection{Spectral Group Encodings}
\label{sec:group_ab}
We hypothesize that, due to the use of sine-cosine positional encodings for spectral groups, the choice of which spectral group encodings to use when fine-tuning is important. Specifically, the model may not be able to leverage the meaning of each spectral group and instead will use the encoding as purely positional. To test this, when fine-tuning with Sentinel-2 alone, we try initializing the spectral group encodings with the encodings corresponding to the spectral group indexing during pre-training (1, 2, 3 where the NAIP spectral group is 0) and compare it with initializing the encodings with encodings corresponding the spectral group indexing during fine-tuning (0, 1, 2). 

We find that using the spectral group encodings corresponding to the position of the sequence, rather than the spectral group itself, leads to performance improvements across most tested settings (Table~\ref{tab:spectral_group_encoding_ab1}) and downstream datasets (Table~\ref{tab:spectral_group_encoding_ab2}). We note that the relative ordering of the performance of USatMAE compared to the baselines is the same with both approaches. This may suggest that the model is using the spectral group encoding to simply distinguish patches from different spectral groups as opposed to leveraging information about the spectral group itself. A learnable spectral group embedding could allow the model to better leverage the spectral group information, but we leave this to future work.

\begin{figure*}[!t]
    \centering
    % \subfigure[Cosine similarity between reference 16x16 patches positional encoding and itself.]{\label{fig:cosine_16}
    % \includegraphics[width=0.49\textwidth]{images/16x16pe.png}}
    % \subfigure[Cosine similarity between reference 16x16 patches positional encoding and 8x8 superpositional encoding.]{\label{fig:cosine_8}
    % \includegraphics[width=0.49
    % \textwidth]{images/8x8pe.png}}
  \begin{subfigure}{.49\linewidth}
    \centering\includegraphics[width=1\linewidth]{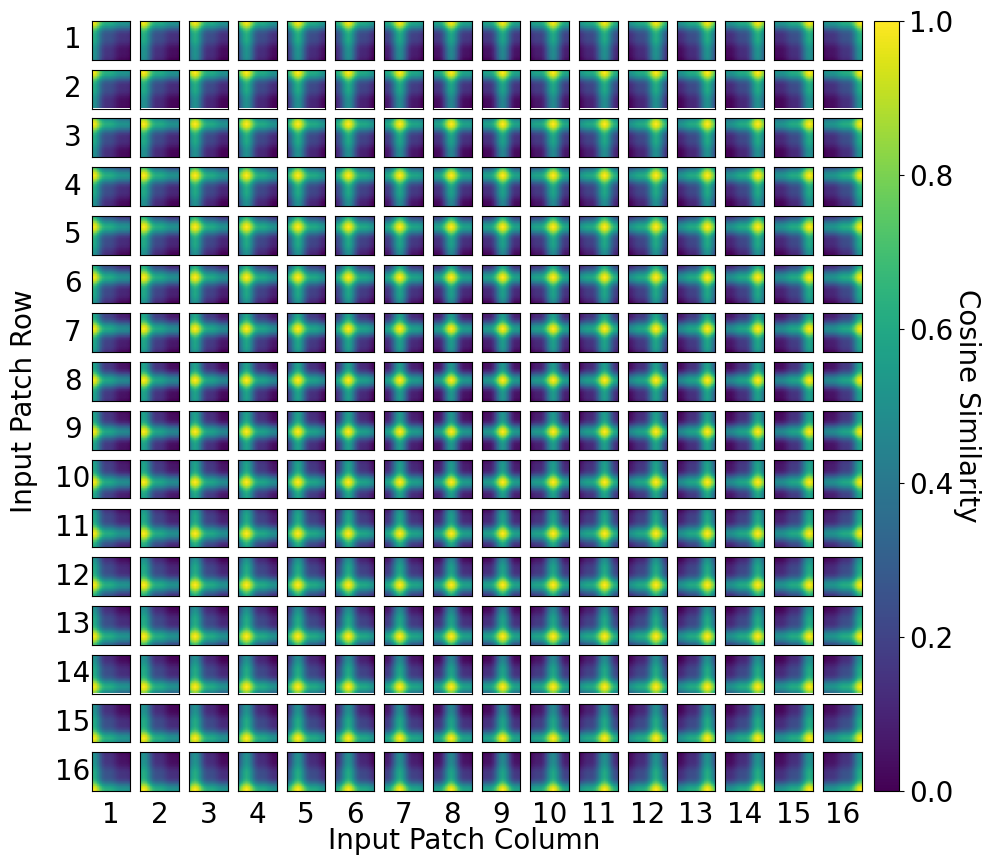}
    \caption{Cosine similarity between reference 16x16 patches positional encoding and itself.}
    \label{fig:cosine_16}
  \end{subfigure}
  \begin{subfigure}{.49\linewidth}
    \centering\includegraphics[width=1\linewidth]{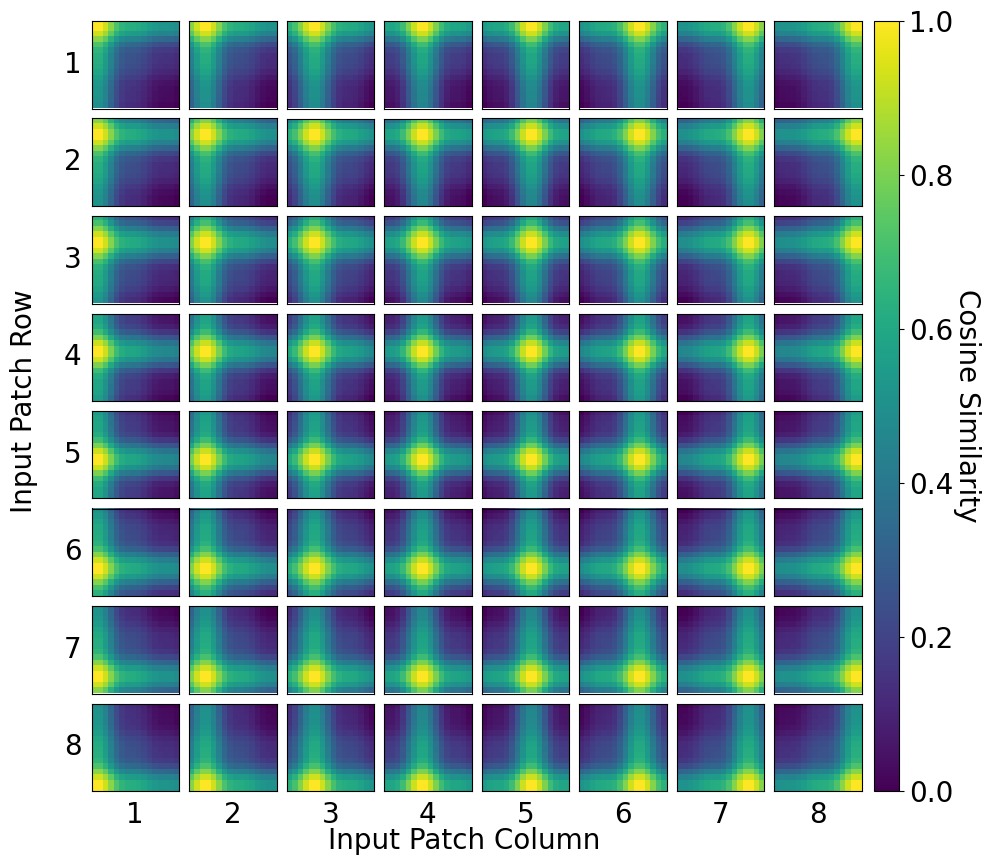}
    \caption{Cosine similarity between reference 16x16 patches positional encoding and 8x8 superpositional encoding.}
    \label{fig:cosine_8}
  \end{subfigure}
    \caption{Visualization of the cosine similarities between positional encodings of an image with 16x16 patches and the superpositional encodings of an image with 8x8 patches.}
    \label{fig:superpositional_sim}
\end{figure*}

\begin{figure*}[!t]
    \centering
    % \subfigure[Training]{\label{fig:usatlas_train}
    % \includegraphics[width=1\textwidth]{images/usatlas_train.png}}
    % \subfigure[Validation]{\label{fig:usatlas_val}
    % \includegraphics[width=1\textwidth]{images/usatlas_val.png}}
  \begin{subfigure}{.49\linewidth}
    \centering\includegraphics[width=1\linewidth]{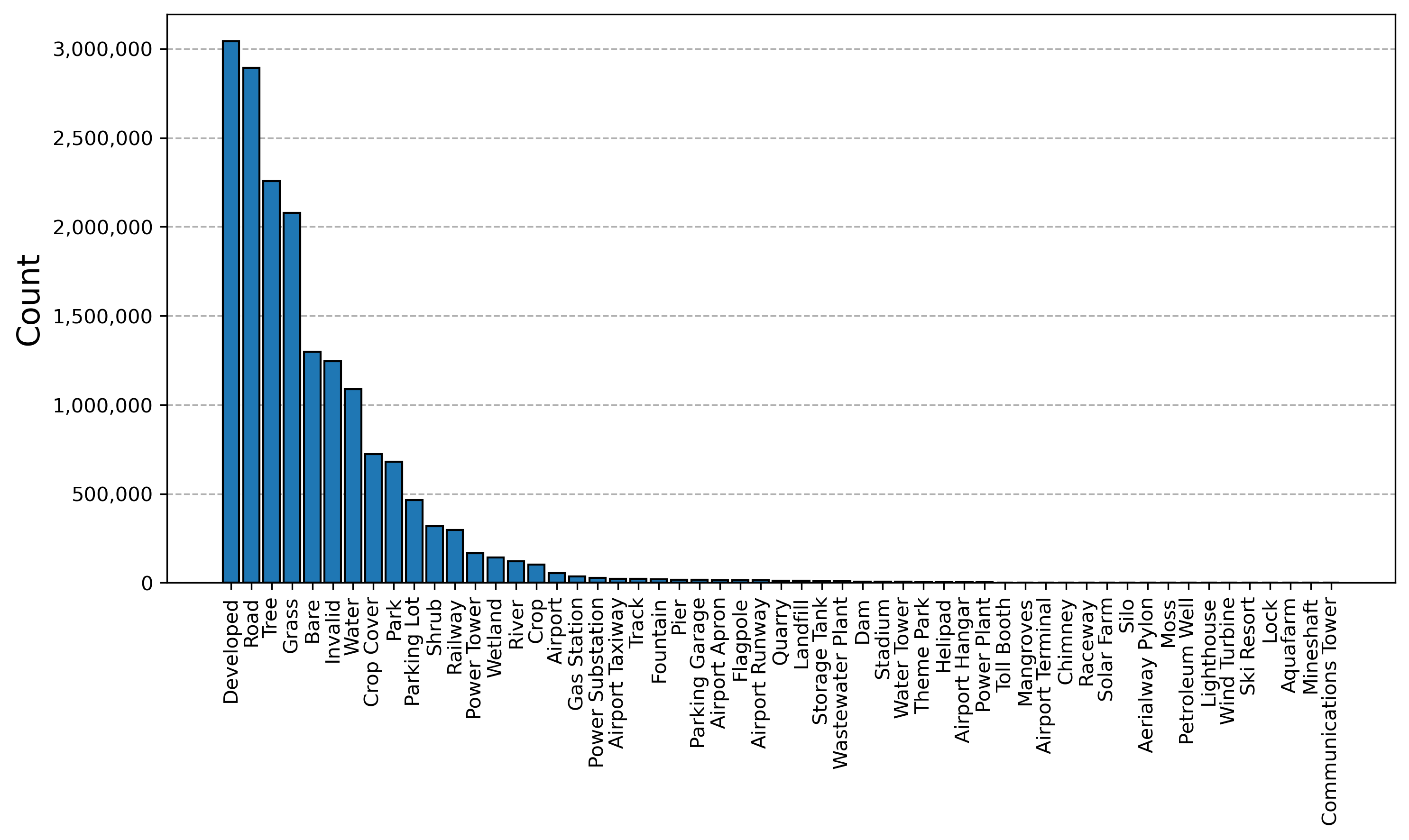}
    \caption{Training}
    \label{fig:usatlas_train}
  \end{subfigure}
  \begin{subfigure}{.49\linewidth}
    \centering\includegraphics[width=1\linewidth]{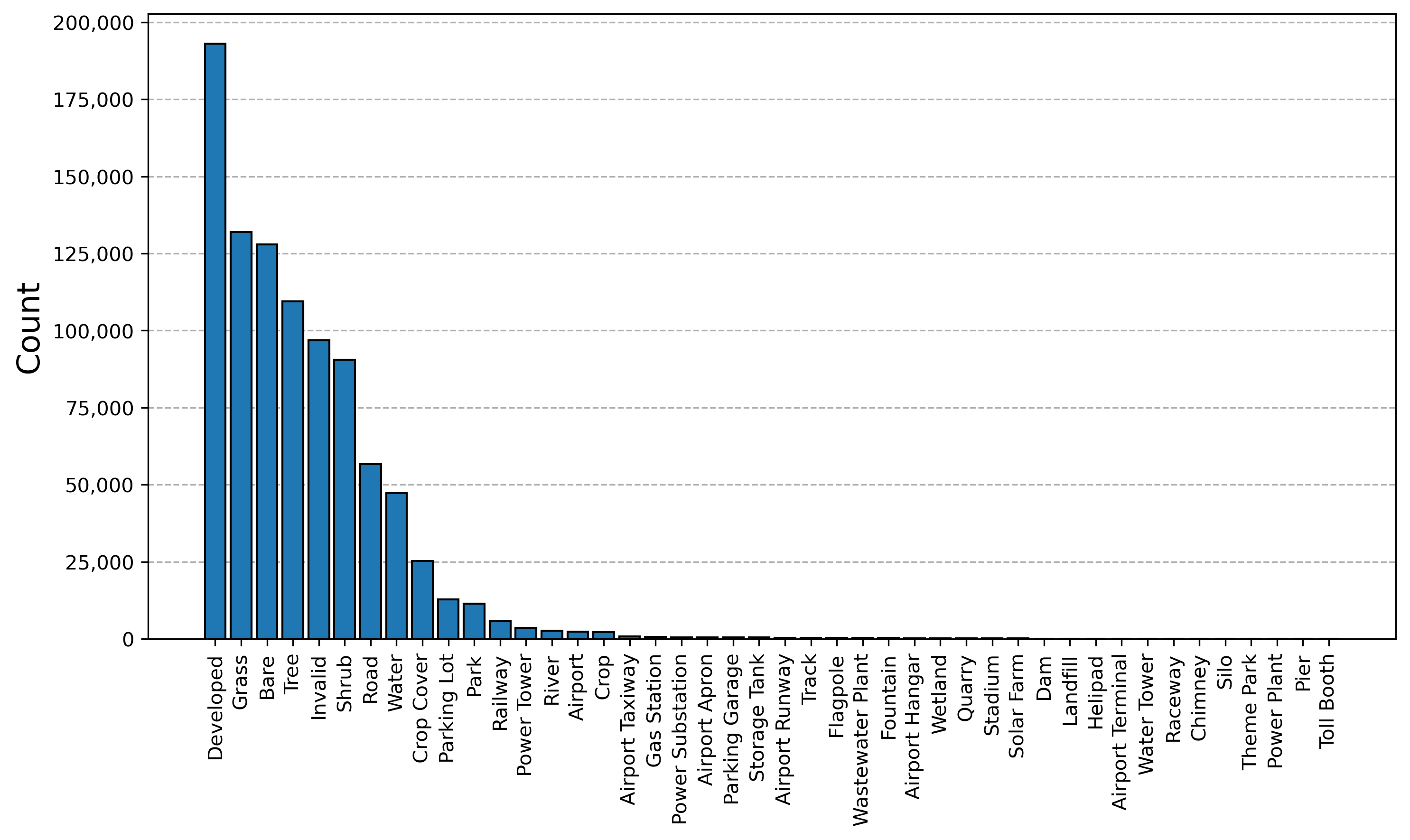}
    \caption{Validation}
    \label{fig:usatlas_val}
  \end{subfigure}
    \caption{Class counts of the USatlas training and validation sets.}
    \label{fig:usatlas}
\end{figure*}

\begin{table}[t!]
    \centering
    \begin{tabular}{c|c}
        \hline
        Spectral Group Pooling & mAP  \\
        \hline
        Sum & 69.97 \\
        Average & 69.78 \\
        \hline
    \end{tabular}
    \caption{USatlas validation set performance of USatMAE using different types of spectral group pooling. Each of these experiments fine-tune using both sensors.}
    \label{tab:pooling}
\end{table}

\begin{table}[t!]
    \centering
    \resizebox{0.6\columnwidth}{!}{%
    \begin{tabular}{c|c|c}
    \hline
        Fine-Tuning Sensor & Spectral Group & mAP \\
        \hline
        Sentinel-2 & - & 66.33\\
        Sentinel-2 & \checkmark & 67.01\\
        NAIP & - & 69.22\\
        NAIP & \checkmark & 69.02\\
        Both & - & 70.37\\
        Both & \checkmark & 69.78\\
        \hline
    \end{tabular}%
    }
    \caption{Ablation testing the effect of spectral group encodings when fine-tuning with different sensors. Performance is measured on the USatlas validation set. Each model is fine-tuned from a model pre-trained using the corresponding encoding settings.}
    \label{tab:sensor_encoding_ab}
\end{table}

\begin{table}[!t]
    \centering
    \resizebox{0.5\columnwidth}{!}{%
    \begin{tabular}{c|c}
        \hline
        Spectral Bands & mAP  \\
        \hline
        Red & 61.99 \\
        Red, Green & 65.53 \\
        Red, Green, Blue & 66.73\\
        Red, Red-Edge 1, SWIR1 & 62.43 \\
        \hline
    \end{tabular}%
    }
    \caption{USatlas validation set performance comparison of USatMAE using different subsets of Sentinel-2 spectral bands.}
    \label{tab:spectral_bands}
\end{table}

\begin{table}[t!]
    \centering
    \resizebox{\columnwidth}{!}{%
    \begin{tabular}{c|c|ccc}
    \hline
        \multirow{2}{*}{Superpositional} & \multirow{2}{*}{Sensor} & \multicolumn{2}{c}{Spectral Group Encoding Index} & \multirow{2}{*}{Diff}\\
        & & Pre-Training & Fine-Tuning  &  \\
        \hline
        - &  - & 65.85 & 66.88 & +1.03\\
        - &  \checkmark & 65.38 & 65.81 & +0.43\\
        \checkmark & - & 67.01 & 66.99 & -0.02\\
        \checkmark & \checkmark & 65.83 & 66.46 & +0.63 \\
        \hline
    \end{tabular}%
    }
    \caption{USatlas validation set performance comparing the effect of using spectral group encodings corresponding to the indices used during pre-training and fine-tuning with different combinations of superpositional and sensor encodings. We pre-train with both sensors and fine-tune with Sentinel-2 for these experiments.}
    \label{tab:spectral_group_encoding_ab1}
\end{table}

\begin{table}[t!]
    \centering
    \resizebox{\columnwidth}{!}{%
    \begin{tabular}{c|ccc}
    \hline
        \multirow{2}{*}{Dataset} & \multicolumn{2}{c}{Spectral Group Encoding Index} & \multirow{2}{*}{Diff}\\
        & Pre-Training & Fine-Tuning &  \\
        \hline
        EuroSAT & 98.02 & 98.37 & +0.35 \\
        BigEarthNet & 85.70 & 85.82 & +0.12 \\
        METER-ML Sentinel-2 & 71.39 & 73.95 & +2.56 \\
        \hline
    \end{tabular}%
    }
    \caption{Downstream performance of multi-sensor pre-trained USatMAE comparing the effect of using spectral group encodings corresponding to the indices used during pre-training and fine-tuning. As in the other downstream experiments, we use superpositional encodings and no sensor encoding for these experiments.}
    \label{tab:spectral_group_encoding_ab2}
\end{table}

% \begin{table}[t!]
%     \centering
%     \begin{tabular}{c|c|c|c}
%     \hline
%         Superpositional & Spectral Group & Masking & mAP \\
%         \hline
%         \checkmark & \checkmark & C & 65.53\\
%         \checkmark & - & C & 64.13\\
%         \checkmark & \checkmark & I & 65.07\\
%         \checkmark & - & I & \textbf{65.83}\\
%         \hline
%     \end{tabular}
%     \caption{Ablation testing the effect of superpositional and spectral group encodings under different masking scehems on USatlas validation set performance with Sentinel-2. C indicates consistent masking between spectral groups and I indicates inconsistent masking between the groups.}
%     \label{tab:masking_ab}
% \end{table}

\begin{figure*}
    \centering
    \includegraphics[width=0.6\linewidth]{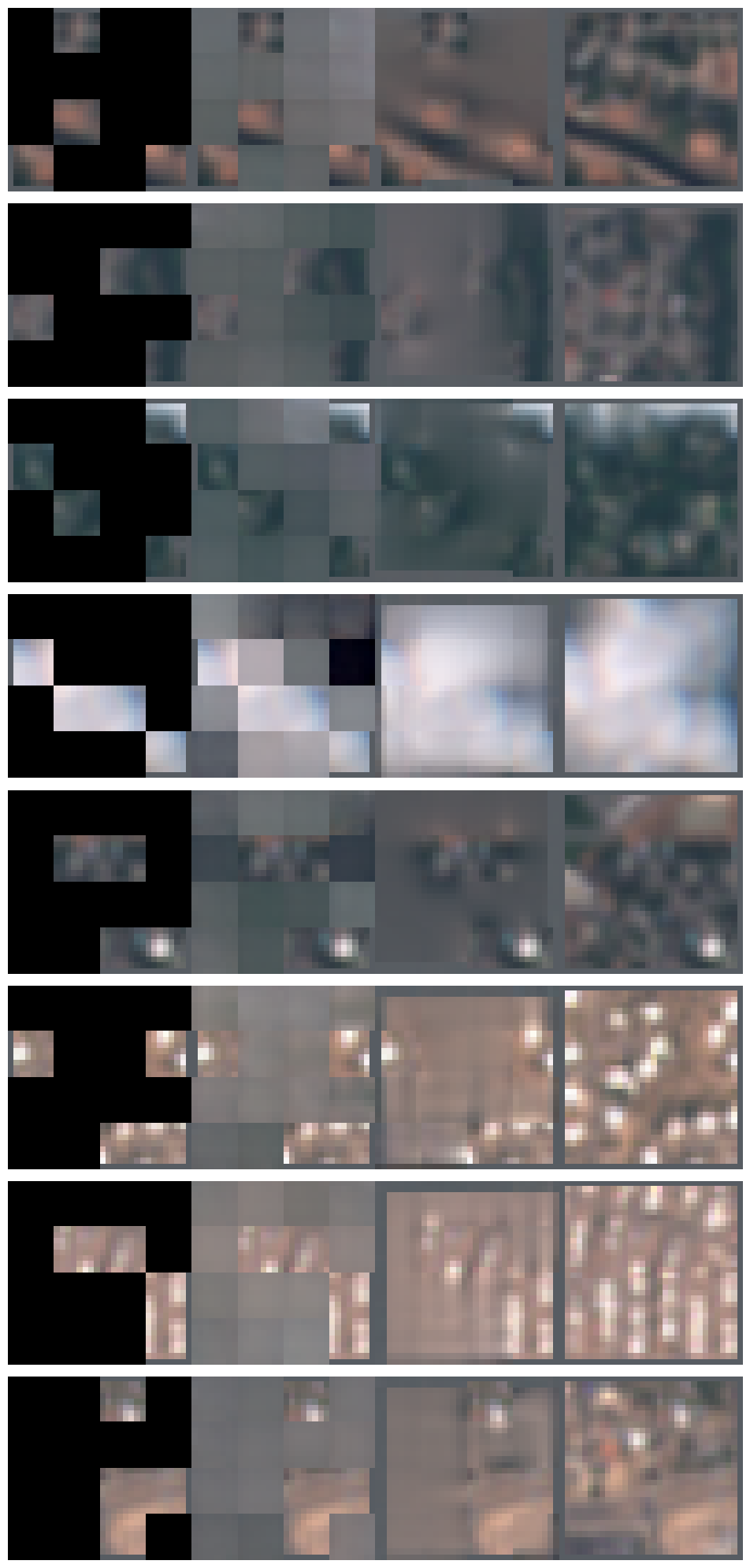}
    \caption{Uncurated random samples of spatially inpainted Satlas Sentinel-2 validation images. We show the masked RGB image (first position), our USatMAE Sentinel-2 RGB reconstruction (second position), our USatMAE Both RGB reconstruction (third position), and the unmasked RGB image (fourth position). The masking ratio is 75\%.}
    \label{fig:inpainting_s2}
\end{figure*}

\begin{figure*}
    \centering
    \includegraphics[width=0.6\linewidth]{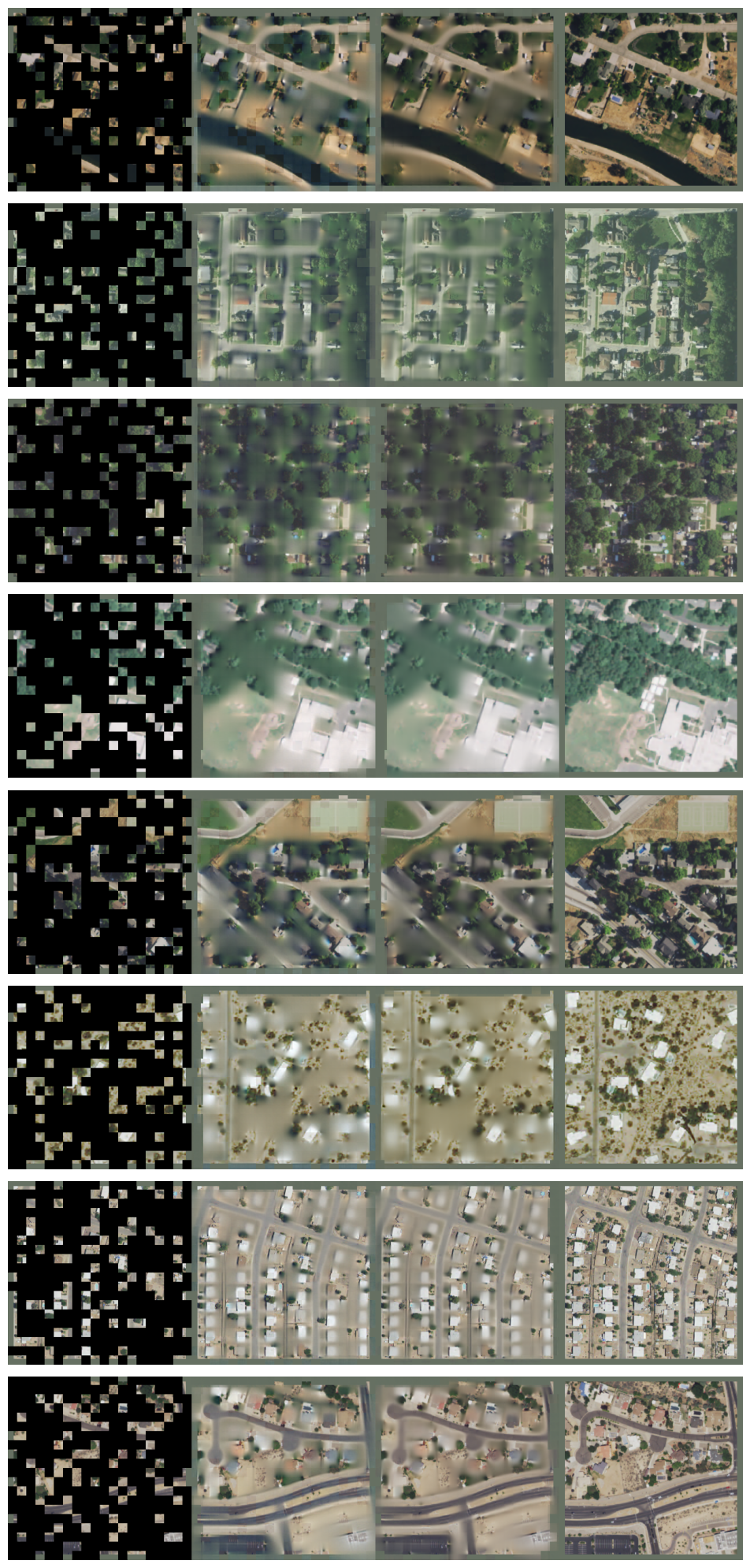}
    \caption{Uncurated random samples of spatially inpainted Satlas NAIP validation images. We show the masked image (first position), our USatMAE NAIP reconstruction (second position), our USatMAE Both reconstruction (third position), and the unmasked image (fourth position). The masking ratio is 75\%.}
    \label{fig:inpainting_naip}
\end{figure*}

% \begin{figure*}
%     \centering
%     \includegraphics[width=0.5\linewidth]{images/mae_reconstruction.png}
%     \caption{Uncurated random samples of spectrally inpainted Satlas NAIP validation images. We mask the green and blue bands and only input the red bands, then task the model with generating the other bands. We show the red band (single channel) image (first position), our USatMAE Both reconstruction of the blue and green bands (second and third positions), and the original blue and green bands (fourth and fifth positions).}
%     \label{fig:inpainting_bands}
% \end{figure*}